\documentclass{article}

\usepackage[preprint]{neurips_2021}

\usepackage[utf8]{inputenc} 
\usepackage[T1]{fontenc}    
\usepackage{hyperref}       
\usepackage{url}            
\usepackage{booktabs}       
\usepackage{amsfonts}       
\usepackage{nicefrac}       
\usepackage{microtype}      
\usepackage{xcolor}         
\usepackage{bm}
\usepackage{graphicx}
\usepackage{amsmath}
\usepackage{subfigure}
\usepackage{caption}
\title{Probabilistic partition of unity networks: \\ clustering based deep approximation}

\newcommand{\MG}[1]{\textcolor{black}{#1}}

\author{%
  Nat Trask \\
  Center for Computing Research \\
  Sandia National Laboratories \\
  Albuquerque, NM 87123 \\
  \texttt{natrask@sandia.gov} \\
   \And
   Mamikon Gulian \\
   Center for Computing Research \\
   Sandia National Laboratories \\
   Albuquerque, NM 87123 \\
   \texttt{mgulian@sandia.gov} \\
   \And
   Andy Huang\\
   Electrical Models and Simulation \\
   Sandia National Laboratories \\
   Albuquerque, NM 87123 \\
   \texttt{ahuang@sandia.gov}\\
   \And
   Kookjin Lee\\
   Quantitative Modeling and Analysis \\
   Sandia National Laboratories \\
   Albuquerque, NM 87123 \\
   \texttt{koolee@sandia.gov}\\
}

\begin{document}

\maketitle

\begin{abstract}
Partition of unity networks (POU-Nets) have been shown capable of realizing algebraic convergence rates for regression and solution of PDEs, but require empirical tuning of training parameters. We enrich POU-Nets with a Gaussian noise model to obtain a probabilistic generalization amenable to gradient-based minimization of a maximum likelihood loss. The resulting architecture provides spatial representations of both noiseless and noisy data as Gaussian mixtures with closed form expressions for variance which provides an estimator of local error. The training process yields remarkably sharp partitions of input space based upon correlation of function values. This classification of training points is amenable to a hierarchical refinement strategy that significantly improves the localization of the regression, allowing for higher-order polynomial approximation to be utilized. The framework scales more favorably to large data sets as compared to Gaussian process regression and allows for spatially varying uncertainty, leveraging the expressive power of deep neural networks while bypassing expensive training associated with other probabilistic deep learning methods. Compared to standard deep neural networks, the framework demonstrates $hp$-convergence without the use of regularizers to tune the localization of partitions. We provide benchmarks quantifying performance in high/low-dimensions, demonstrating that convergence rates depend only on the latent dimension of data within high-dimensional space. \MG{Finally, we} introduce a new open-source data set of PDE-based simulations of a semiconductor device and perform unsupervised extraction of a \MG{physically interpretable reduced-order basis}.
\end{abstract}

\section{Introduction}
\begin{figure}[t]
    \centering
    \includegraphics[width=0.99\textwidth]{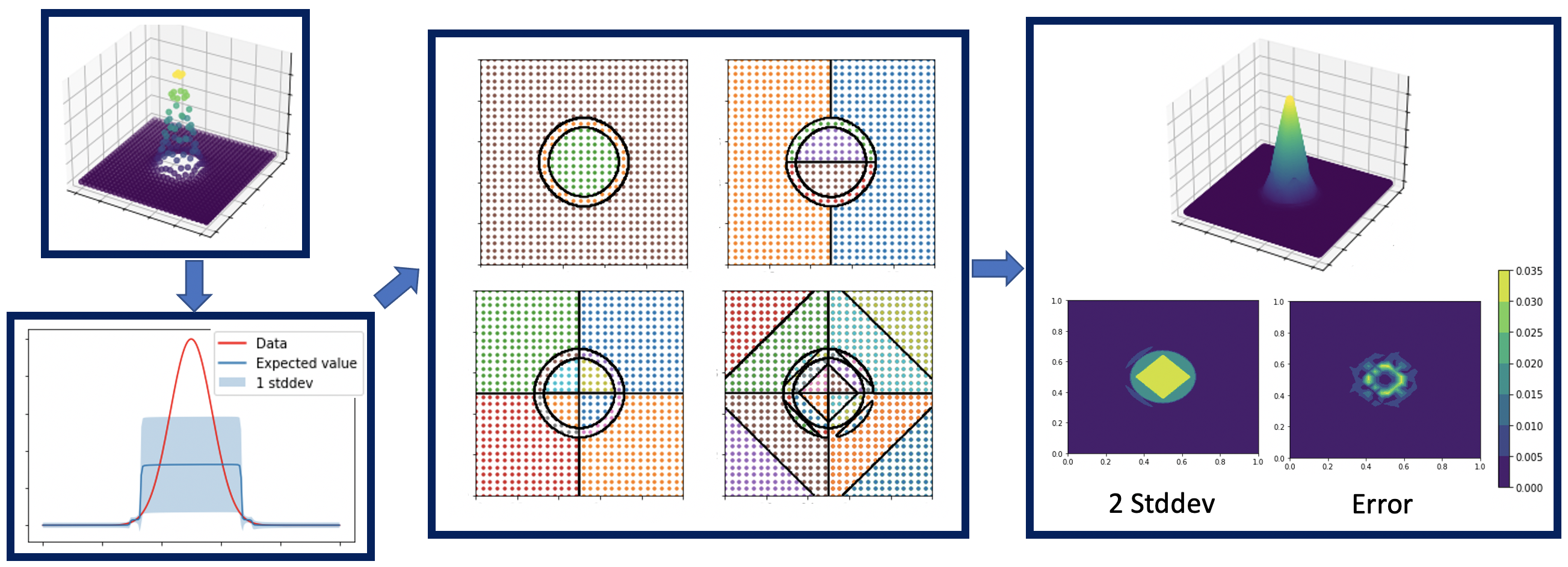}
    \caption{Flowchart of the PPOU-Net training process presented in this paper. (1) Data is sampled from a deterministic or noisy function at unstructured locations. (2) A partition of unity network (POU-Net) with univariate Gaussian noise is fit to data using a maximum likelihood estimate (MLE). The training of the means and variances clusters the training data into three sets based on their $y$-values. (3) The resulting partitions are generally sharp but not simply connected. Repeated bisection via principle component analysis provides simply connected partitions of space. (4) With compactly supported partitions, polynomials are fit to partitions in a least squares sense, and the MLE is updated. Closed form expressions for the variance of the PPOU provide a local error estimator with respect to training data.}
    \label{fig:masterFig}
\end{figure}
The construction of approximation spaces is central to scientific computation. For the vast majority of partial differential equations (PDEs)\MG{,} simulations of physical systems rely upon the generation of a mesh and a corresponding set of shape functions \citep{johnson2012numerical,braess2007finite}. Generally\MG{,} a mesh partitions space into disjoint subvolumes to which one may attach local polynomial spaces to construct a finite element method (FEM). The mesh generation process is labor intensive, requiring expert human intervention to avoid degenerate partitions and obtain desirable approximation properties; \MG{\citet{boggs2005dart} reveals that up to 60\% of person-hours spent by engineers on design-to-solution is spent on mesh generation.} The automated construction of approximation spaces, particularly to process high-throughput geometric datasets, is \MG{therefore} an attractive target for machine learning (ML). 

During training, multilayer perceptrons (MLPs) construct data-driven bases with no reference to underlying geometry \citep{cyr2020robust,he2018relu}. 
It has been proven that for MLPs of increasing width and depth, weights and biases exist for which the Sobolev norm of approximation error converge algebraically. 
In theory, this suggests MLPs may be competitive with traditional FEM \citep{opschoor2019deep,he2018relu}.
In practice, however, the training error of MLPs optimized with first-order methods fails to converge at all in large architecture limits \citep{fokina2019growing}. Partition-of-unity networks (POU-nets) \citep{lee2021partition} repurpose classification architectures to instead partition space and obtain localized polynomial approximation, and have recently demonstrated $hp$-convergence during training. 
Successful training to obtain compactly supported partitions appropriate for polynomial approximation required empirical tuning of a regularizer, and we aim to develop a parameter-free approach.

In this work we introduce probabilistic partition of unity networks (PPOU-Nets) by enriching the polynomial spaces in POU-Nets with a noise model (See \ref{fig:masterFig}). This implictly regularizes the learning problem and allows a robust means of partitioning space without the need for penalty parameters. The partition of unity underpinning POU-Nets admits interpretation as a probability distribution, allowing interpretation of noise as a Gaussian mixture which may be used in concert with polynomial least squares regression. Fitting the maximum likelihood of the PPOU-Net model provides an unsupervised means of partitioning space with first-order optimizers. We will demonstrate this has value both in performing regression with noiseless and noisy data while estimating uncertainty, establishing algebraic convergence independent of ambient dimension. Finally, we conclude with an application extracting a piecewise constant reduced-order basis encoding a large, high-resolution dataset of semiconductor responses. Unlike traditional principal component analysis (PCA)-based reduced order models (ROMs) \citep{hesthaven2016certified,quarteroni2014reduced}, the resulting basis is tied to a volumetric partition of space, a property which has been shown to be effective for extracting structure-preserving data-driven models \citep{trask2020enforcing}.

\noindent\textbf{Relation to previous work:}
As discussed below, our method is based on an architecture which resembles a tensor product of a Gaussian mixture model in output space and a POU-Net in input space. 
Gaussian mixture models (GMMs) have been combined with neural network architectures in many classification models. Some approaches integrate GMMs into outputs or hidden layers of the architecture \citep{variani2015gaussian,tuske2015integrating} of the architecture, while others involve GMM distributions for parameters \citep{viroli2019deep}. Other applications include speech recognition \citep{lei2013deep} or rating prediction \citep{deng2018neural}. Training probabilistic neural networks using likelihood loss functions has also been studied for classification problems \citep{streit1994maximum,perlovsky1991maximum,jia2021forecasting}. Some architectures have been proposed that are analagous in that context to the approach we consider for regression \citep{tsuji1995maximum,perlovsky1994model,traven1991neural}.
To our knowledge, this approach has not been studied for regression with the POU-Net architecture we utilize. 

Many models have attempted to incorporate Bayesian elements into deep learning (see e.g. Bayesian networks \citep{mackay1992practical,jospin2020hands}, dropout recovering deep Gaussian process \citep{gal2016dropout}, others \citep{osawa2019practical}). Often, these models require costly Monte Carlo training strategies, while the current approach can be trained with gradient descent. The current approach adopts the probabilistic viewpoint with the aim of improving approximation, while the previously cited works generally focus on quantifying uncertainty. In a deterministic context several works have pursued other strategies to realize convergence in \MG{deep networks \citep{he2018relu,cyr2020robust,adcock2021gap,fokina2019growing,ainsworth2021galerkin}}. In the context of ML for reduced basis construction, several works have focused primarily on using either Gaussian processes and PCA \citep{guo2018reduced} or classical/variational autoencoders as replacements for PCA \citep{lee2020model,lopez2020variational} in classical ROM schemes; this is distinct from the control volume type surrogates considered in \citep{trask2020enforcing} which requires a reduced basis corresponding to a partition of space.

\section{Method}
We consider the problem of regressing a function $y(x)$ from scattered data $\mathcal{D} = \left\{x_i,y_i\right\}_{i=1}^N$, where $x_i\in \Omega \subset \mathbb{R}^d$. We will also consider the case where $x_i$ are sampled from a manifold $\mathcal{M}\subset\mathbb{R}^d$ with latent dimension \MG{$d_l = \text{dim}(\mathcal{M}) < d$.} A partition of unity (POU) is defined as a collection of $M$ nonnegative functions $\phi_i:\Omega\rightarrow\mathbb{R}$ satisfying $\sum^M_i \phi_i(x) = 1$ for all $x$. In POU-Nets, the set of POUs is parameterized as a deep neural network with a final softmax layer; we denote $\Phi_{\theta_1} := \left\{\phi_i(x;\theta_1)\right\}_{i\in\mathbb{R}^M}$, where $\theta_1$ are the hidden layer parameters. For concreteness, we will exclusively consider a depth five residual network architecture initialized with the Box initializer \citep{cyr2020robust}. For each partition we attach a function $p_i \in V$, where $V$ is a Banach space; in this work we adopt the $m{\text{th}}$-order polynomials $V=\pi_m(\mathbb{R}^d)$. The POU-Net architecture \citep{lee2021partition} is then defined as
\begin{equation}\label{eq:pounet_original}
    y_{\text{pou}}(x;\theta_1) = \sum_i \phi_i(x;\theta_1) p_i(x).
\end{equation}
This architecture admits an explicit least-squares solution for minimizer of a mean square loss, 
exactly reproduces data $y \in V$, and converges algebraically with respect to $m$ \MG{if $\phi_i$ have compact support}. To obtain localized partitions, \citet{lee2021partition} required empirical regularizers on $p_i$.

The POU-Net architecture \eqref{eq:pounet_original} has a natural probabilistic interpretation, given that POUs admit interpretation as discrete probability densities. For fixed $x$, one \MG{samples} a one-hot vector $\bm{\Phi}(x,\xi;\theta_1)$, with probability $\phi_i(x;\theta_1)$ of realizing a one in the $i{\text{th}}$ entry and $\xi$ indexing the sample. Then
\begin{equation}
y_{\text{pou}}(x;\theta_1) = \underset{\xi}{\mathbb{E}} \left[\sum_i {\Phi}_i(x,\xi;\theta_1) p_i(x)\right].
\end{equation}
Now, we associate to each partition $i$ a random variable $X_i(\theta_2)$ representing additive noise  whose distribution is characterized by parameters $\theta_2$. \MG{We} define an analogous sample of the probabilistic partition on unity network (PPOU-Net) model as
\begin{equation}\label{eq:PPOU_sample}
    y_{\text{ppou}}(x,\xi,\omega;\theta_1,\theta_2) = \sum_i {\Phi}_i(x,\xi;\theta_1) \big(p_i(x) + X_i(\omega,\theta_2)\big),
\end{equation}
where $X_i(\omega,\theta_2)$ denotes a sample of $X_i(\theta_2)$ \MG{with sample index $\omega$.}
 This leads to the following generative model: for fixed $x$, one selects partition $i$ with discrete probability $\phi_i$ and then samples $\left(p_i(x) + X_i(\omega;\theta_2)\right)$. It is clear that in the absence of the additive noise ($X_i=0$ for all $i$) we recover the standard POU-Net \MG{generative model}.

For this work we choose to use univariate Gaussians as noise model\MG{, i.e. $X_i \sim \mathcal{N}(y;\mu_i,\sigma_i)$}, denoting $\theta_2 = \left\{\mu_i,\sigma_i\right\}_i$. In this special case, for fixed $x$\MG{, the PPOU-Net} reduces to learning a Gaussian mixture model \citep{reynolds2009gaussian} \MG{expressed as} combinations of Gaussians weighted by the POU. Denote the evaluation of the deterministic component \MG{by}
\begin{equation}
    Q(x) = \sum_{i=1}^M \phi_i(x; \theta_1) p_i(x).
\end{equation}
Then the probability density function \MG{$p(y_{\text{ppou}}(x; \theta_1, \theta_2))$ governing the distribution of \eqref{eq:PPOU_sample} is given by}
\begin{equation}\label{eq:y_density_2}
\MG{p(y_{\text{ppou}}(x;\theta_1, \theta_2)) = 
\sum_{i=1}^M \phi_i(x; \theta_1) \mathcal{N}\left(
y_{\text{ppou}}(x) \, \big| \, \mu_i + Q(x) , \sigma_i\right),}
\end{equation}
\MG{where $\mathcal{N}(y \,| \, \mu, \sigma)$ denotes the normal density with mean $\mu$ and standard deviation $\sigma$}.
\MG{The mean $\mu_y$ and variance $\sigma_y$ of the random variable $y_{\text{ppou}}(x)$ are then given explicitly by}
\begin{align}
\label{eq:mean_prediction}
\mu\MG{_y}(x) &= \sum_{i=1}^M \phi_i(x; \theta_1) (\mu_i + Q (x)),
\\
\label{eq:st_dev_prediction}
\sigma_y(x) &= 
\sum_{i=1}^M \phi_i(x; \theta_1) \sigma_i^2
+
\sum_{i=1}^M \phi_i(x; \theta_1) \mu_i^2
-
\left( \sum_{i=1}^M \phi_i(x; \theta_1) \mu_i \right)^2.
\end{align}
\MG{Denoting by $\mathcal{N}(\mu, \sigma)$ the random variable with density $\mathcal{N}(y \,| \, \mu, \sigma)$, equation \eqref{eq:y_density_2} implies that
\begin{equation}
y_{\text{ppou}}(x; \theta_1, \theta_2) \sim Q(x) + 
\sum_{i=1}^M \phi_i(x; \theta_1) \mathcal{N}(\mu_i, \sigma_i),
\end{equation}
which represents $y_{\text{ppou}}$ as a POU-Net prediction augmented by Gaussian mixture uncertainty.    
}

\subsection{Multivariate Density and Likelihood}
For the joint statistics for $y(x_1), y(x_2), ..., y(x_N)$, we assume independence
to obtain the multivariate density
\begin{align}
p(y(x_1), y(x_2), ..., y(x_N)) = \prod_{\ell=1}^N p(y(x_\ell)) =
\prod_{\ell=1}^N
\sum_{i=1}^M \phi_i(x_\ell; \theta_1) \mathcal{N}(y(x_\ell); \mu_i + Q (x_\ell) , \sigma_i),
\end{align}
where $p(y(x_\ell))$ is the density given by \eqref{eq:y_density_2}.
Given data $\mathcal{D}$, we can evaluate the above density and treat it as a likelihood. 
We \MG{then} define a likelihood loss
\begin{align}
\mathcal{L}(\MG{\theta_1, \bm{\mu}, \bm{\sigma}}) 
&= 
-\log\left( p(y(x_1), y(x_2), ..., y(x_N))  \right) \\
&=
-\log\left( \prod_{\ell=1}^N
\sum_{i=1}^M \phi_i(x_\ell; \theta_1) \mathcal{N}(y(x_\ell); \mu_i + Q (x_\ell) , \sigma_i)  \right) \\
&=
-\sum_{\ell=1}^N \log \left(
\sum_{i=1}^M \phi_i(x_\ell; \theta_1) \mathcal{N}(y(x_\ell); \mu_i + Q (x_\ell) , \sigma_i) \right).
\end{align}
Here, $\bm{\mu}$ and $\bm{\sigma}$ are vectors with components $\mu_i$ and $\sigma_i$; below, we refer to $\{\bm{\mu},\bm{\sigma}\}$ as $\theta_2$. 
We minimize this loss using the Adam algorithm \citep{kingma2014adam}. For all examples, a learning rate of $0.01$ is used to emphasize the lack of sensitivity to model parameters.

\subsection{Least squares polynomial fit}
The parameters defining $Q$, i.e., the coefficients of the polynomials corresponding to each partition, can be updated by minimizing the likelihood loss. An alternative step to obtain an estimate of $Q$ is to solve the following least squares problem for fixed $\theta_1$,
\begin{equation}
    p_i^*(x;\theta_1
    ) = \underset{p_i\in \pi_{m}}{\text{argmin}} \sum_{j=1}^N \sum_{i=1}^M \left( \phi_i(x_j; \theta_1) (p_i(x_j) - y(x_j) \right)^2,
\end{equation}
and define $Q^*_m = \sum_{i=1}^M \phi_i(x; \theta_1) p^*_i(x).$ Note that this least square estimator amounts to solving a linear system of size $M \text{dim}(\pi_{m})$.

\subsection{PCA-based bisection of partitions}\label{subsec:PCA}
When training PPOU-Nets, we observed that the optimizer partitions the $y$-axis by placing the $\mu_i$ \MG{and} selecting the $\sigma_i$ to cluster \MG{the} $y$-values of the data. This effects a \MG{``soft''} classification of \MG{the data} into $M$ nearly-disjoint classes or intervals, \MG{labelled by the POU index $i$.} The optimizer concurrently partitions space during training by adapting the supports of the POU functions $\phi_i(x; \theta_1)$ to the boundaries of the $x$-values of the data that have been \MG{so labelled}. We found that this training is excellent at producing nearly disjoint partition of domain space \MG{by the sets $\text{supp}(\phi_i)$} without explicit regularization. However, it has the undesirably property that the sets $\text{supp}(\phi_i)$ are not simply connected; this is visible in the examples in Section \ref{sec:1d_examples}.
To post-process partitions one could use any segmentation algorithm (e.g. \MG{$k$}-means \citep{likas2003global}, connected components \citep{hopcroft1973algorithm}, etc). For simplicity, after the \MG{$\phi_i$} are trained, we pursue a principal component analysis (PCA) based bisection strategy which reliably yields simply-connected, quasi-uniform partitions in a post-processing step.
Given a collection of points $\mathcal{D}_i \subset \mathcal{D}$, define
\begin{equation}
    C_i = \sum_{x_j \in \mathcal{D}_i} \left(x_j - \bar{x}_i\right)\otimes\left(x_j - \bar{x}_i\right),
\end{equation}
where $\bar{x}_i = \MG{(\#\mathcal{D}_i)^{-1}} \sum_{x_j \in \mathcal{D}_i} x_j$ is the center of mass. Performing the singular value decomposition of $C_i$ provides a best fit of an ellipsoid to $\mathcal{D}_i$, with the axis direction given by the right singular vectors and the lengths given by the corresponding singular values. We denote the vector corresponding to the largest singular value $\bm{n}_i$. 

The function $F(x) = \text{argmax} \{ \phi_i(x) \}$ classifies the data according into $M$ classes. Taking $\mathcal{D}_i = F^{-1}(i)$, we obtain two new partitions $\phi_{i,+}\MG{(x)} = \phi_i(x)*\mathbf{1}_{(x-\bar{x}_i)\cdot \bm{n}_i > 0}\MG{(x)}$ and $\phi_{i,-}\MG{(x)} = \phi_i(x)*\mathbf{1}_{(x-\bar{x}_i)\cdot \bm{n}_i \MG{\leq} 0}\MG{(x)}$, where $\mathbf{1}_A$ denotes the indicator function of a set $A$. Note that this decomposition preserves the POU property because $\phi_i = \phi_{i,+} + \phi_{i,-}$. In practice\MG{,} this may allow for a hierarchical refinement which may be performed adaptively - for this work\MG{,} however\MG{,} we will consider a fixed number of refinements $N_{\text{ref}}$. This yields a total of $M_{\text{tot}} = M \times 2^{N_{\text{ref}}}$ partitions of input space. 

\section{Experiments}
\subsection{1D deterministic and noisy regression}\label{sec:1d_examples}

\begin{figure}[t]
    \centering
    \includegraphics[width=0.32\textwidth]{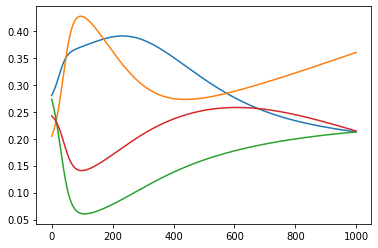}
    \includegraphics[width=0.32\textwidth]{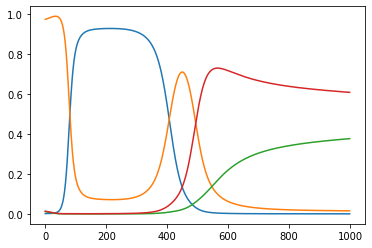}
    \includegraphics[width=0.32\textwidth]{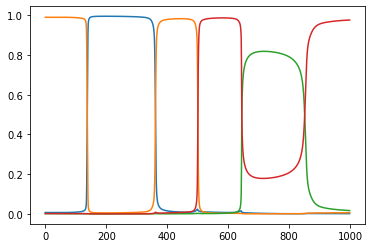}\\
    \includegraphics[width=0.32\textwidth]{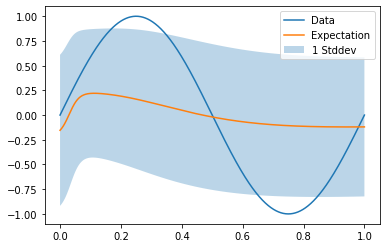}
    \includegraphics[width=0.32\textwidth]{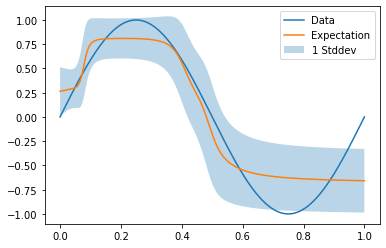}
    \includegraphics[width=0.32\textwidth]{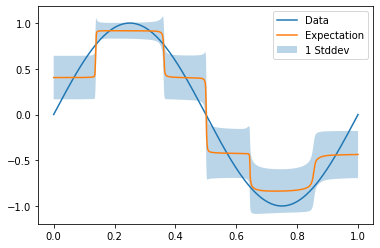}\\
    \includegraphics[width=0.32\textwidth]{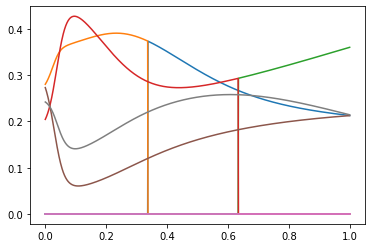}
    \includegraphics[width=0.32\textwidth]{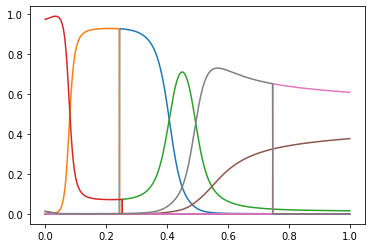}
    \includegraphics[width=0.32\textwidth]{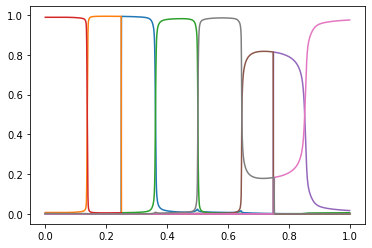}\\
    \includegraphics[width=0.32\textwidth]{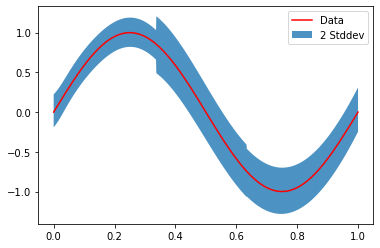}
    \includegraphics[width=0.32\textwidth]{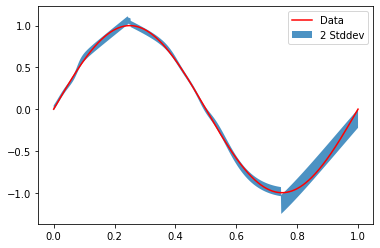}
    \includegraphics[width=0.32\textwidth]{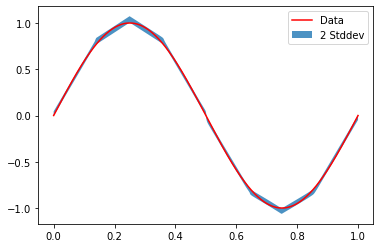}
    \caption{Evolution of PPOU \MG{at initialization} (\textit{left column}), 1\MG{,}000 steps (\textit{center column}) and 10\MG{,}000 steps (\textit{right column}). \textit{Row 1:} initially random partitions evolve into approximation of indicator functions. \textit{Row 2:} Training of MLE without polynomial contribution $Q$ yields piecewise constant expectation on each partition with standard deviation estimating error. \textit{Row 3:} Bisection provides simply connected partitions amenable to polynomial approximation. \textit{Row 4:} Piecewise polynomial regression with an estimate of uncertainty with no human in the loop.}
    \label{fig:cleanData}
\end{figure}

We \MG{first} consider regression of the function $y = \sin 2 \pi x$\MG{, illustrated in} in Figure \ref{fig:cleanData}. To train the model, we first apply 10\MG{,}000 iterations of Adam to $\underset{\theta_1,\bm{\mu},\bm{\sigma}}{\text{min}}\mathcal{L}(\theta_1, \bm{\mu}, \bm{\sigma}, 0)$. This provides \MG{an initial} partition of space consistent with clustering the range of data on the $y$-axis and approximating indicator functions on sets of functions $x$-realizing those values. The \MG{mean prediction \eqref{eq:mean_prediction}} provides an approximately piecewise constant representation of the data, and the standard deviation provides an estimate of the piecewise error on each partition. Secondly, to obtain simply connected partitions we apply a single refinement of PCA bisection (\ref{subsec:PCA}) and then assign $Q = Q^*_1$ to obtain a piecewise linear fit to the bisected partitions. Thirdly, we apply 500 iterations of Adam to $\underset{\bm{\sigma}}{\text{min}} \, \mathcal{L}(\theta_1,\MG{\bm{\mu} = \bm{0}}, \bm{\sigma}, Q^*_1)$ to obtain uncertainties regarding the fit polynomials\MG{; note that $\bm{\mu}$ is redundant with constant terms in $Q_1^*$.} For the remainder of the work, we follow this same three-step approach, varying the number $N_{\text{ref}}$ of bisections but keeping the other \MG{training} parameters fixed. 

\begin{figure}[]
    \centering
    \begin{minipage}{.65\textwidth}
    \includegraphics[width=0.49\textwidth]{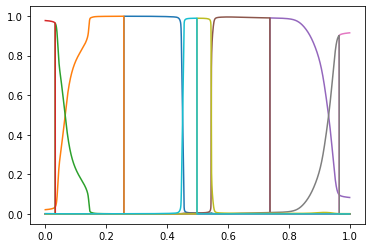}
    \includegraphics[width=0.49\textwidth]{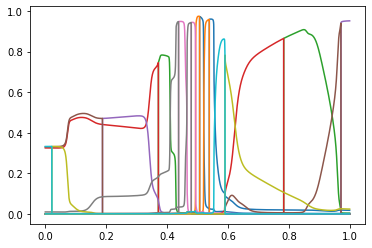}\\
    \includegraphics[width=0.49\textwidth]{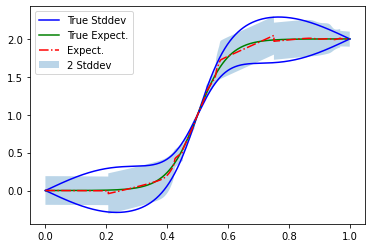}
    \includegraphics[width=0.49\textwidth]{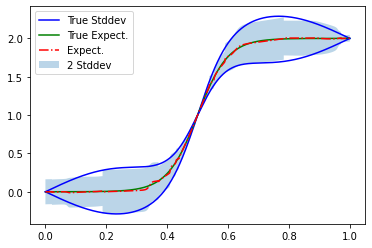}
    \end{minipage}
    \begin{minipage}{0.33\textwidth}
    \includegraphics[width=\textwidth]{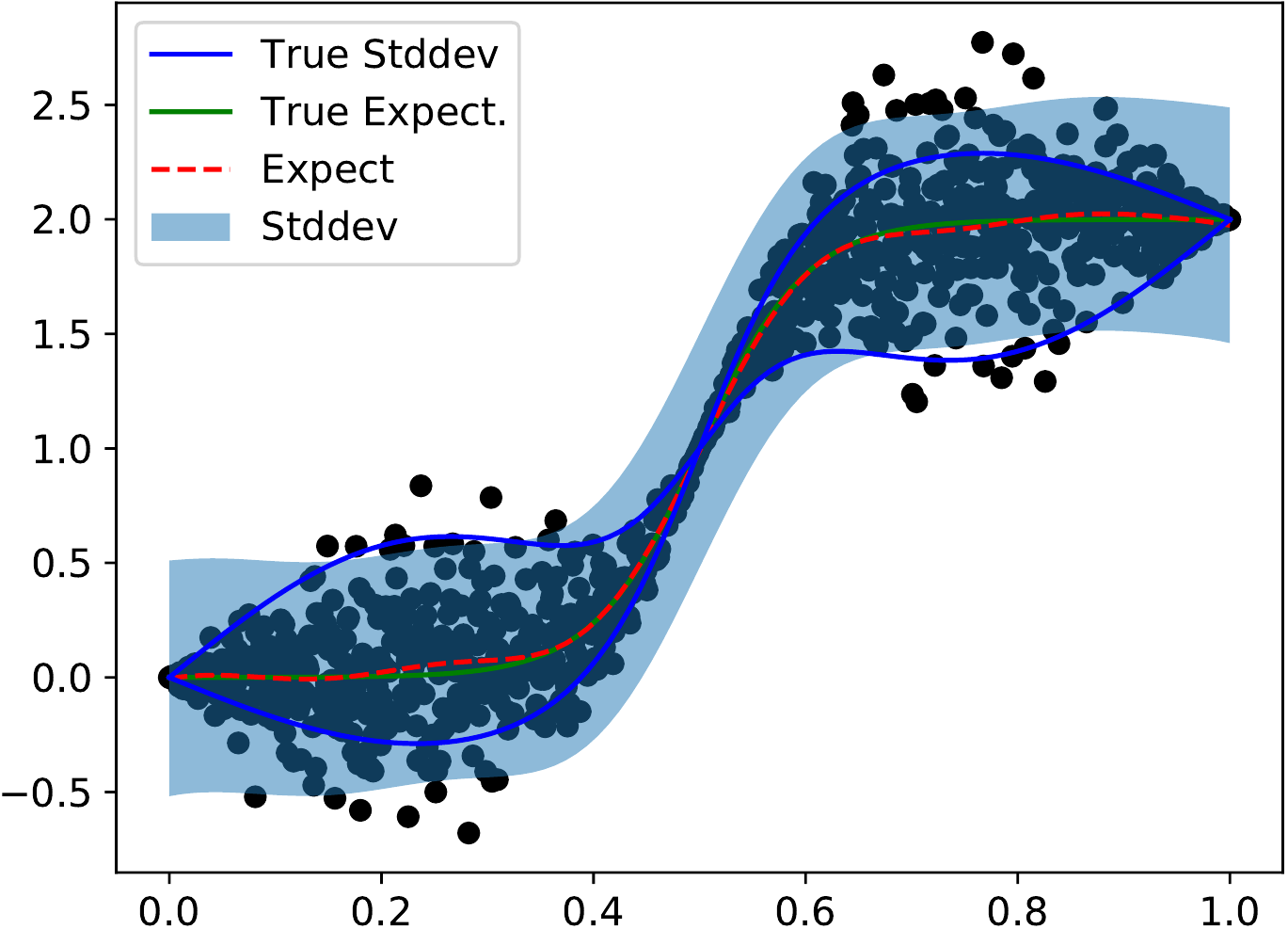}
    \end{minipage}
    \caption{Regression of smooth function with spatially varying noise using \MG{$M=5$} partitions \textit{(left column)} and \MG{$M=10$} partitions \textit{(center column)}. Partitions \textit{(top)} cluster automatically near steep gradients in either function value or uncertainty to accurately estimate standard deviation. In comparison, Gaussian process regression \textit{(right column)} is unable to provide an accurate estimate of uncertainty, since the constant noise amplitude $\sigma_{GP}$ is unable to model the spatially heterogeneous noise.}
    \label{fig:noisyData}
\end{figure}

We next consider regression of functions with spatially heterogeneous white noise
\begin{equation}
y = 1 + \tanh{10(x-0.5)} + \epsilon, \quad \epsilon \sim \mathcal{N}(0,\sigma = 0.3 \sin(2 \pi x) ),
\end{equation}
\MG{as illustrated in Figure \ref{fig:noisyData}.}
We train identically to the previous example, generating  1,000 samples spaced evenly on $[0,1]$ using \MG{$M=5$ and $M=10$} partitions. PPOU-Nets \MG{balance} placement of partitions against noise in the data and gradients in the underlying function, clustering partitions at the sharp gradient in the middle and \MG{tightening} variance at the endpoints to accurately denoise the signal. By comparison, a typical Gaussian process regression with a Gaussian likelihood, i.e., a covariance matrix of the form $k(X,X') + \sigma_{\text{GP}}^2 \text{Id}$\MG{,} is unable to provide a spatially varying characterization of uncertainty when using a stationary square-exponential covariance kernel function $k$. The difference in computational cost in this large data example is significant: PPOU-Nets perform the regression in minutes, while the the Guassian process regression (implemented using \texttt{scikit-learn} \MG{\citep{scikit-learn}}) requires more than an hour, due to the cubic scaling of covariance matrix inversion in the number of observations.

\subsection{Breaking the curse-of-dimensionality: Dependence of accuracy on dimension}

\begin{figure}[]
\raisebox{-0.5\height}{\begin{minipage}{.70\textwidth}
    \centering
    \includegraphics[width=0.32\textwidth]{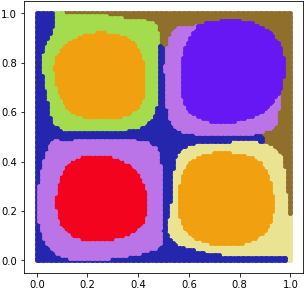}
    \includegraphics[width=0.32\textwidth]{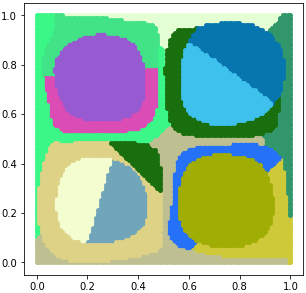}
    \includegraphics[width=0.32\textwidth]{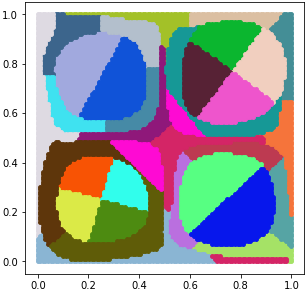}\\
    \includegraphics[width=0.32\textwidth]{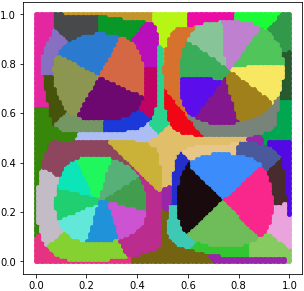}
    \includegraphics[width=0.32\textwidth]{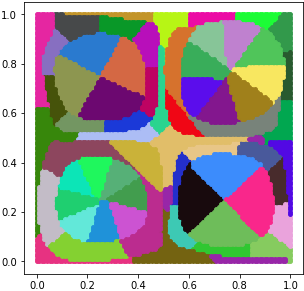}
    \includegraphics[width=0.32\textwidth]{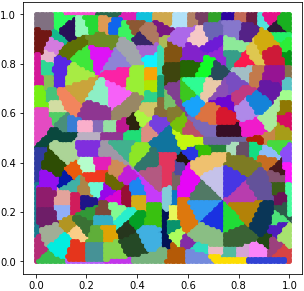}\\
\end{minipage}}
\raisebox{-0.5\height}{\begin{minipage}{.29\textwidth}
    \includegraphics[width=\textwidth]{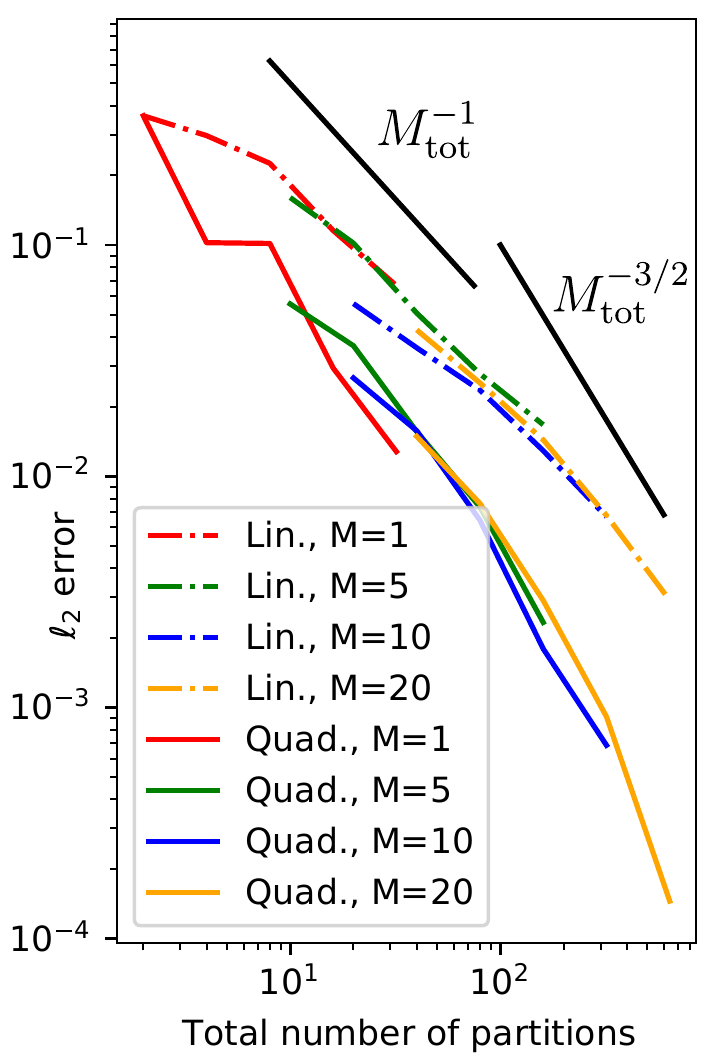}
    \end{minipage}}
    \caption{Regressing $y=\sin 2 \pi x \sin 2 \pi y$ in two dimensions \textit{(left)} provides partitions of more complex topology. Bisection is critical to obtain compactly supported sets amenable to polynomial approximation, while initial partition focuses refinement. Comparison of linear and quadratic regression for $M = 1,5,10,20$ initial partitions demonstrate algebraic convergence rates under refinement consistent with FEM approximation theory.}
    \label{fig:lowdimSins}
\end{figure}

In this section we consider the convergence rate of the PPOU-Nets\MG{,} its interaction with the ambient dimension $d$ and, in the case where $\mathcal{D}$ is sampled from a latent low dimensional manifold \MG{$\mathcal{M}$}, its interaction with \MG{$d_l = \text{dim}(\mathcal{M})$}. Standard approximation results for $m{\text{th}}$-order polynomial based approximation on quasi-uniform compact partitions suggest a scaling of root-mean-square error $\epsilon \sim M_{\text{tot}}^{-\frac{m+1}{d}}$, for total number of partitions $M_{\text{tot}}$, polynomial order $m$, and relevant space dimension $d$ \citep{wendland2004scattered}. Figure \ref{fig:lowdimSins} illustrates representative partitions in two dimensions as well as the desired convergence rates for linear ($m=1$) and quadratic ($m=2$) polynomials. 

\begin{figure}[t]
    \centering
    \raisebox{-0.5\height}{\includegraphics[width=0.37\textwidth]{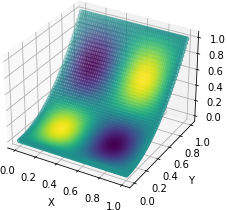}}
    \raisebox{-0.5\height}{\includegraphics[width=0.33\textwidth]{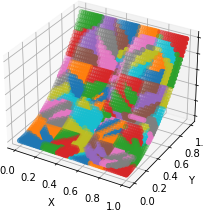}}
    \raisebox{-0.5\height}{\includegraphics[width=0.28\textwidth]{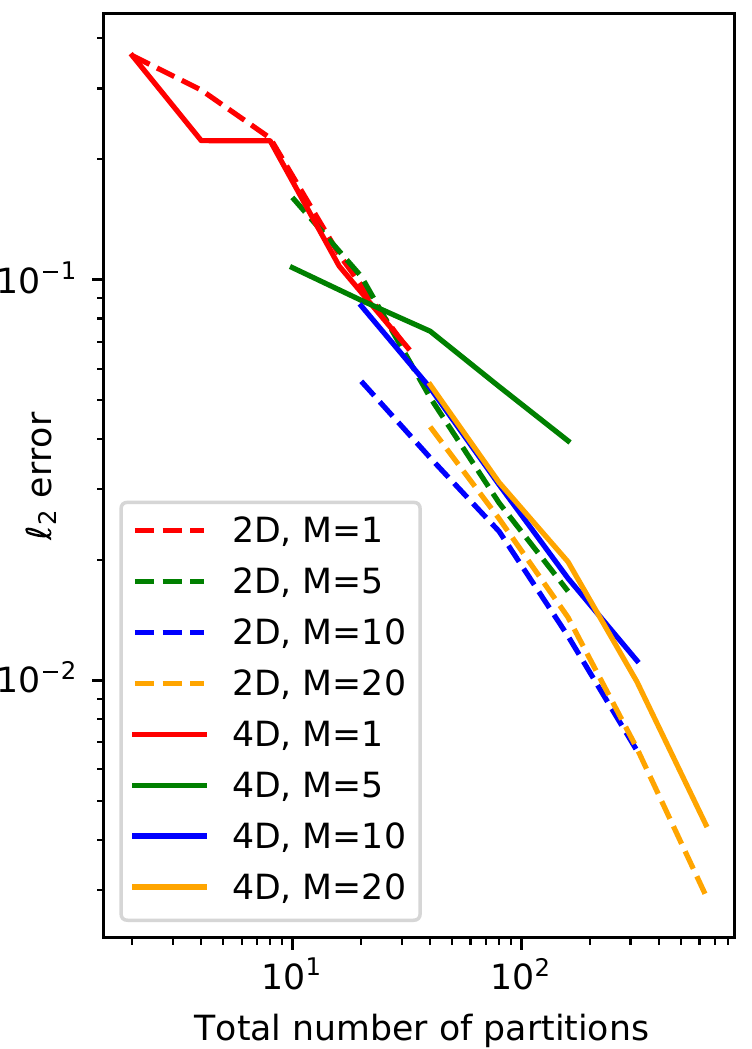}}
    \caption{Lifting the data from Fig \ref{fig:lowdimSins} by mapping $(x_1,x_2) \in \mathbb{R}^2$ to $(x_1,x_2,x_2^2,0) \in \mathbb{R}^4$  \textit{(left)} provides qualitatively similar partitions of space \textit{(center)} which provide similar convergence rates when comparing the two- and four-dimensional settings \textit{(right)}. Plots depict the projection of data $(x_1,x_2,x_3,x_4) \in \mathbb{R}^4$ into $(x_1,x_2,x_3,0) \in \mathbb{R}^3$.}
    \label{fig:highdimSins}
\end{figure}

In Figure \ref{fig:highdimSins} we embed this two-dimensional dataset into four dimensions by mapping points $(x_1,x_2) \in \mathbb{R}^2$ to $(x_1,x_2,x_2^2,0) \in \mathbb{R}^4$. The input data is therefore four dimensional, though lying on a latent manifold of dimension $d_l = 2$.
Training with linear polynomials ($m=1$), we observe the the resulting partitions exhibit similar patterns, while the same convergence rates for the error are observed as in two dimensions for linear polynomials (reproduced from Figure \ref{fig:lowdimSins} for comparison). This suggests that the well-documented ability of MLPs to break the "curse-of-dimensionality" allows discovery of meshes whose approximation power scales with $d_l$ rather than $d$. Moreover, the PCA-based refinement strategy built on these meshes also scales according to $d_l$ rather than $d$. 

\subsection{Application: partitioning of semiconductor data}

\begin{figure}[t]
    \centering
    \raisebox{-0.5\height}{\includegraphics[width=0.9\textwidth]{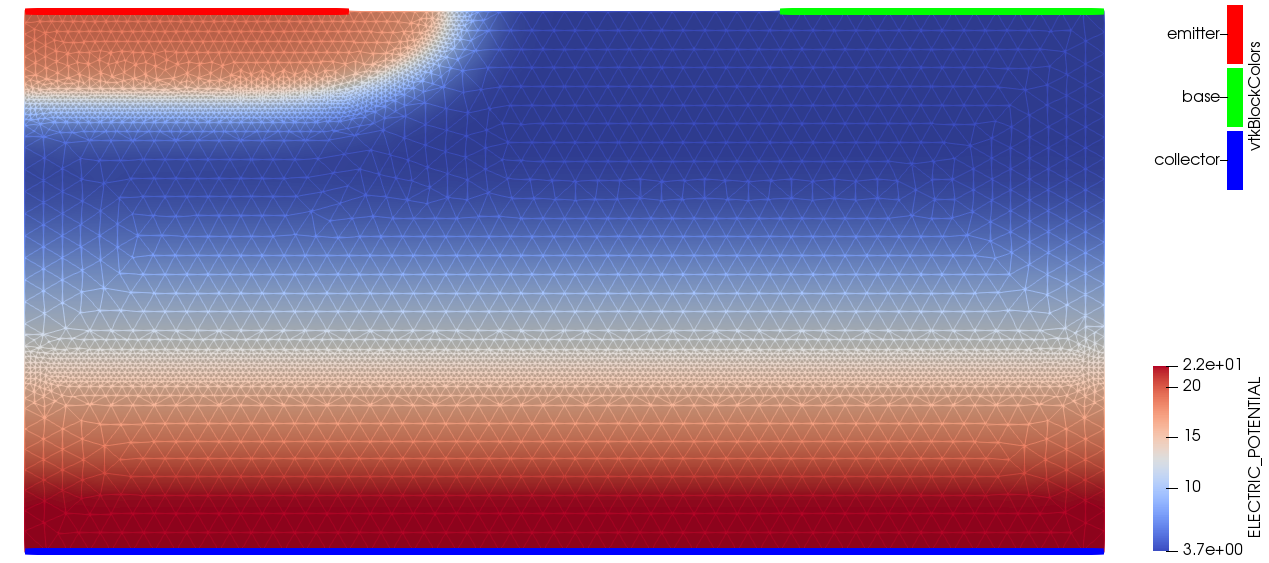}}\\
    \raisebox{-0.5\height}{\includegraphics[width=0.49\textwidth]{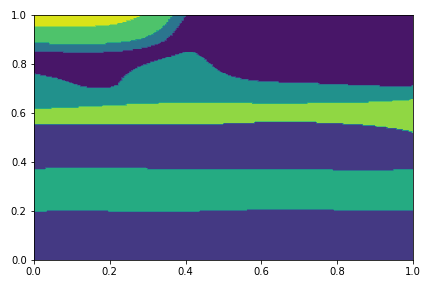}}
  \raisebox{-0.5\height}{\includegraphics[width=0.49\textwidth]{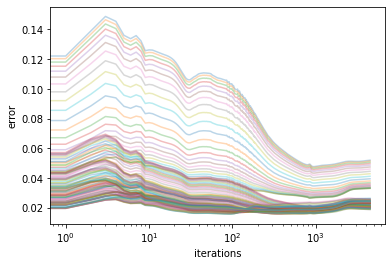}}
    \caption{\textit{(Top:)} A representative solution from the BJT database demonstrating the electric potential field on a finite element grid with 4,154 degrees of freedom (basis functions). The potential is the solution of the boundary value problem obtained by imposing a fixed voltage $V_1$ at the emitter (top left red boundary), $V_2$ at the base (top right green boundary), and $V_3$ at the collector (bottom blue boundary). The database is constructed by performing a three dimensional sweep over $(V_1,V_2,V_3)$. \textit{(Bottom left:)} Applying the PPOU process with $m=0$ and $N_{\text{ref}}=0$ provides an estimate for the best set of 10 partitions which approximate any given PDE solution in the database well. The location of partitions coincide with physical intuition - the sharp gradients near ``junctions'' at the emitter and baseplate are the locations where the BJT provides ideal diode-like behavior. \textit{(Bottom right:)} Evolution of RMS error for least squares fit of each solution in the database as partitions evolve during training. A worst case error of $<5\%$ is achieved.}
    \label{fig:charon}
\end{figure}

Finally, we apply the process to generate a partition of space design to provide optimal representation of a large, high-resolution database of PDE solutions. We consider 121 solutions to the nonlinear drift-diffusion equations governing behavior of a bipolar junction transistor (BJT) with three terminals. In the semiconductor industry, so-called ``compact models'' provide reduced-order descriptions of devices in terms of simplified algebraic-differential models stemming from Kirchoff's laws. Despite the important role of such models as an interface between circuit designers and device technology \citep{saha2015compact}, their construction generally occurs over timescales of decades for a given device (see, e.g. \citet{liu1998bsim3v3,liu2011bsim4,chauhan2015finfet}). Recent works have sought to use use graph neural networks on partitions of space to automate the extraction of such models \citep{gao2020physics,trask2020enforcing}. To aid in the machine learned construction of compact models, we provide this database of PDE solutions generated using the open-sourced technology computer aided design (TCAD) software library Charon \citep{musson2018charon} (available at https://charon.sandia.gov/).

The probabilistic underpinnings of PPOUs allow the learning of a single partitioning of space which optimally approximates the entire database. For the $k$th solution, $1 \le k \le 121$, we obtain a dataset $\mathcal{D}_k$ with coordinates $\bm{x}_k$ corresponding to the nodes of a finite element mesh, and $\bm{y}_k$ corresponding to the electric potential associated with a given distribution of voltages at the three terminals of the device. By concatenating the labels into $\bm{x} = [\bm{x}_1,...,\bm{x}_{121}]$, $\bm{y} = [\bm{y}_1,...,\bm{y}_{121}]$ we obtain a scattered dataset $\mathcal{D}=\left\{\bm{x},\bm{y}\right\}$ whose uncertainty at a given spatial location corresponds to the spread of solutions in the database. Applying PPOU-Net regression with piecewise constant approximation ($M=10,m=0,N_{\text{ref}}=0$) yields a partitioning of space which minimizes the variance across the data set  (Figure \ref{fig:charon}), and therefore indirectly minimizes the error in representing any single PDE solution in the database with a worst-case error of $5\%$ with a $400\times$ reduction in degrees of freedom. Further, the resulting partitions correspond to parts of the semiconductor device with physical interpretation. This provides an alternative partitioning-based reduced space representation of solution to the typical PCA based approaches used in ROMs which may map better into data-driven modeling contexts such as those studied by \citet{gao2020physics} and \citet{trask2020enforcing}.

\section{Conclusion}
We have presented a novel probabilistic extension of partition of unity networks which includes a model for noise in the output on each partition. During training, the model performs an unsupervised partitioning of space to provide a clustering-based means of approximation, and may be trained simply with gradient-based optimizers. We have demonstrated $hp-$convergence with respect to the number of partitions dependent only upon the latent dimension of data. In the context of quantifying uncertainty in noisy data the architecture appears well-suited for adaptively handling heterogeneous noise. An application to semiconductor data shows that the framework may be well-suited to incorporate into reduced-order models.

In training our PPOU-Net approximation, we have performed both bisection and polynomial fitting to partitions as a postprocessing step after an initial training of the partition of unity functions. In future work, we aim to simultaneously optimize over all parameters $(\theta_1,\theta_2,Q)$. A wide range of architectural choices and refinement strategies may also be explored: e.g. autoencoders for hidden layer architectures, \MG{$k$}-means for partitioning, etc. For the construction of reduced-order bases following \citet{gao2020physics,trask2020enforcing}, we will also explore whether physics constraints can be imposed strongly simultaneously with learning of partitions.


\section{Acknowledgements}
Sandia National Laboratories is a multimission laboratory managed and operated by National Technology and Engineering Solutions of Sandia, LLC, a wholly owned subsidiary of Honeywell International, Inc., for the U.S. Department of Energy’s National Nuclear Security Administration under contract {DE-NA0003530}.  This paper describes objective technical results and analysis.  Any subjective views or opinions that might be expressed in the paper do not necessarily represent the views of the U.S. Department of Energy or the United States Government. 
\MG{SAND Number: {SAND2021-7709 O}.}

The work of N. Trask, and M. Gulian is supported by the U.S. Department of Energy, Office of Advanced Scientific Computing Research under the Collaboratory on Mathematics and Physics-Informed Learning Machines for Multiscale and Multiphysics Problems (PhILMs) project. N. Trask and K. Lee are supported by the Department of Energy early career program. M. Gulian is supported by the John von Neumann fellowship at Sandia National Laboratories. 
The work of A. Huang is supported by the Laboratory Directed Research and Development (LDRD) program, Project No. 218474, at Sandia National Laboratories.

\bibliographystyle{abbrvnat} 
\bibliography{SNL_neurIPS}

\end{document}